\documentclass[conference]{IEEEtran}
\IEEEoverridecommandlockouts
\usepackage{cite}
\usepackage{amsmath,amssymb,amsfonts}
\usepackage{algorithmic}
\usepackage{float}
\usepackage{graphicx}
\usepackage{textcomp}
\usepackage{xcolor}
\usepackage{comment}
\def\BibTeX{{\rm B\kern-.05em{\sc i\kern-.025em b}\kern-.08em
    T\kern-.1667em\lower.7ex\hbox{E}\kern-.125emX}}

\usepackage{caption}
\usepackage[acronym]{glossaries}
\makeglossaries
\usepackage{url}
\newacronym{NGD}{NGD}{natural gas demand}
\newacronym{LDZ}{LDZ}{residential gas demand}
\newacronym{IND}{IND}{industrial gas demand}
\newacronym{GTP}{GTP}{gas to power generation demand}
\newacronym{HDD}{HDD}{heating degree days}
\newacronym{ANN}{ANN}{artificial neural network}
\newacronym{RNN}{RNN}{recurrent neural network}
\newacronym{LSTM}{LSTM}{long-short-term memory}

\begin{document}

\title{Deep Causal Learning to Explain and Quantify The Geo-Tension's Impact on Natural Gas Market}

\author{
\IEEEauthorblockN{Philipp K. Peter$^*$}
\IEEEauthorblockA{\textit{Information Systems} \\ 
\textit{University of Cologne}\\
Cologne, Germany \\
0000-0002-5590-0482}
\and
\IEEEauthorblockN{Yulin Li$^*$}
\IEEEauthorblockA{\textit{Institute of Energy Economics} \\
\textit{University of Cologne}\\
Cologne, Germany \\
liy7@smail.uni-koeln.de
}
\and
\IEEEauthorblockN{Ziyue Li $^\dagger$}
\IEEEauthorblockA{\textit{Information Systems} \\ 
\textit{University of Cologne}\\
Cologne, Germany \\
0000-0003-4983-9352}
\and
\IEEEauthorblockN{Wolfgang Ketter}
\IEEEauthorblockA{\textit{ Information Systems} \\ 
\textit{University of Cologne}\\
Cologne, Germany \\
0000-0001-9008-142X}
\thanks{$^*$ Equal contribution; $^\dagger$ Corresponding Author (zlibn@wiso.uni-koeln.de)}
}

\maketitle

\begin{abstract}
   Natural gas demand is a crucial factor for predicting natural gas prices and thus has a direct influence on the power system. However, existing methods face challenges in assessing the impact of shocks, such as the outbreak of the Russian-Ukrainian war. In this context, we apply deep neural network-based Granger causality to identify important drivers of natural gas demand. Furthermore, the resulting dependencies are used to construct a counterfactual case without the outbreak of the war, providing a quantifiable estimate of the overall effect of the shock on various German energy sectors. The code and dataset are available at \url{https://github.com/bonaldli/CausalEnergy}.
\end{abstract}


\section{Introduction}
The Russia-Ukraine conflict has introduced significant uncertainties in the German and European natural gas markets. Understanding the causal elements behind demand fluctuations and accurately quantifying the conflict's impact is crucial for future energy planning and strategy.  
It is intuitive to formulate such a task as a prediction problem, i.e., the predict the post-conflict energy price. Traditionally, forecasting in this area leaned heavily on linear regression models, time series methods, and other statistical techniques \cite{gorucu2004evaluation}. 
While these conventional methodologies can have their strengths, they often exhibit limitations, especially when dealing with non-linear relationships (especially non-stationarity from the conflict) and multiple influencing factors simultaneously. 
Recently, deep learning models have excelled in analyzing vast amounts of information efficiently, capturing nonlinear relationships, and unearthing patterns and correlations that might be missed by traditional statistical methods \cite{liu2021natural}. 
Beyond prediction, another more important topic will be analyzing \textit{whether} and \textit{how much} the conflict has impacted the energy market. 
Causal analysis offers a foundational framework to answer such a question. Applying these two approaches in combination can be especially valuable. 
In this line of thought, the following two research questions are investigated:

\textbf{RQ1}: Generally, which factors can best describe the patterns of various German energy sectors (including residential, industrial, and gas power plant) so that, ideally, we could use them for better prediction?

\textbf{RQ2}: How does the war affect German energy sectors differently, and is it possible to explain and even quantify these impacts in the long run?

\begin{figure*}[t]
    \centering
    \includegraphics[width=0.99\textwidth]{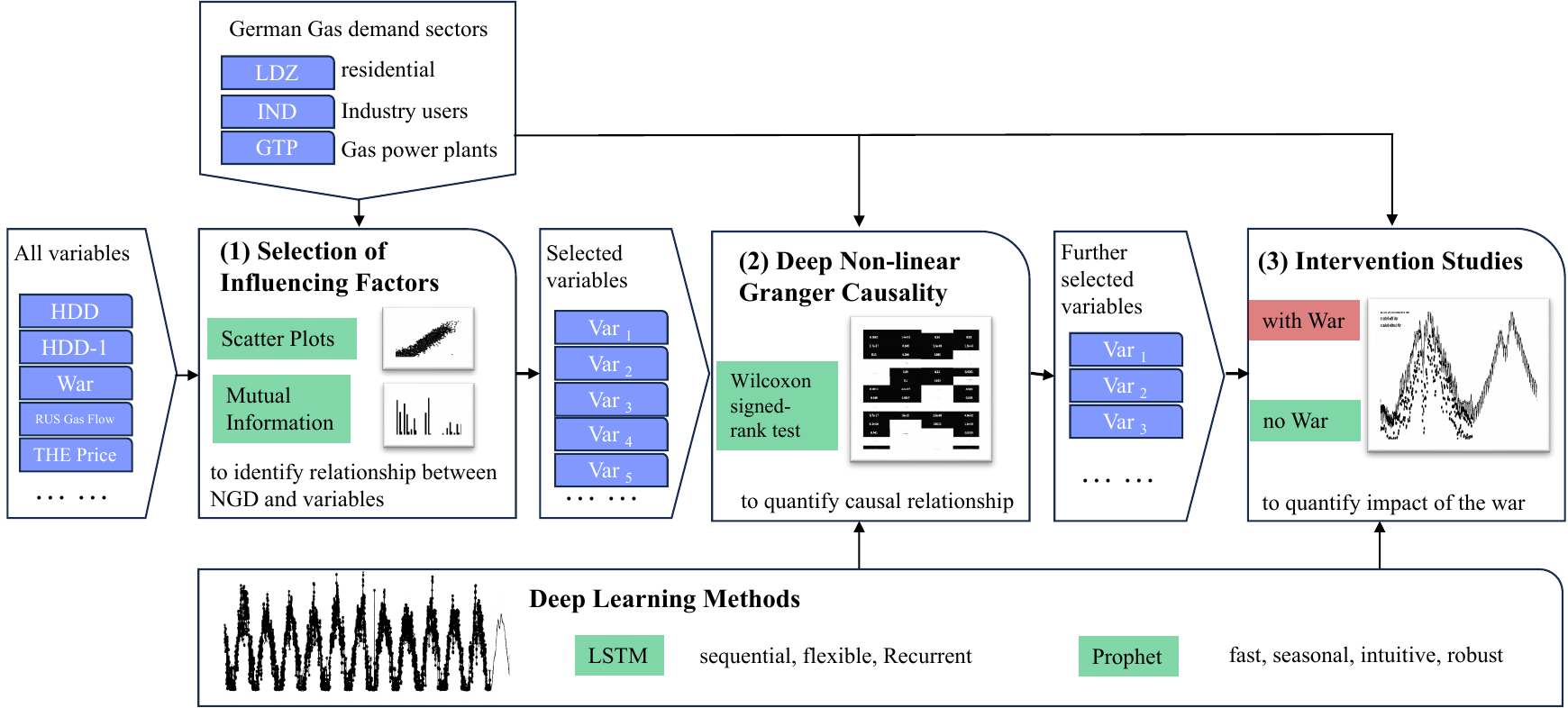}
    \caption{(1) Selection of influencing factors for each sector of LDZ, IND, and GTP; (2) Deep Non-linear Granger Causality based on LSTM and Prophet; (3) Intervention studies based on counterfactual prediction using only pre-war training data.}
    \label{fig:Researchapproach}
    \vspace{-10pt}
\end{figure*}

To address these challenges, this paper adopts a three-pronged approach, with an overview in Fig. \ref{fig:Researchapproach}. Firstly, the different sets of data are investigated based on mutual information, to tailor different selection of factors for different sector. Then, non-linear Granger causality tests \cite{granger1969investigating} are applied based on \acrfull{LSTM} networks \cite{hochreiter1997long}, to identify which factors exhibit a strong influence. Thirdly, an LSTM and a prophet network \cite{taylor2018forecasting} based prediction is generated as a counterfactual scenario for the war period to estimate the effect of the war on natural gas demand.

This paper offers three main contributions to the extant research on forecasting energy demand. First, it proposes a simple and accurate prediction model using deep learning tools for the natural gas demand that is facing unprecedented issues in Germany. Second, the study reveals and quantifies the causal effect of war on future natural gas demand, including residential, industrial, and gas power plant sectors in Germany. Finally, when there are significant shocks to the system, like wars or other unexpected events, the proposed models can be swiftly adapted to different natural gas demand forecasting areas based on the updated information.

The rest of the paper is organized as follows. Section \ref{sec:lr} reviews earlier research on forecasting methods of energy demand. Section \ref{sec:ra} introduces the research approach. Section \ref{sec: result} shows the results, including the selection of influencing factors, nonlinear Granger causality analysis, and intervention studies regarding the effect of war on German natural gas demand across different sectors. 
Section \ref{sec: con} gives the conclusion.

\section{Related Work}
\label{sec:lr}
\subsection{Prediction Methods for Natural Gas Demand}
There has been abundant research in time series prediction based on various techniques such as statistical models \cite{jiang2023unified}, tensor \cite{li2020tensor,tsung2020discussion,li2020long,li2022profile,sergin2024low,yan2024sparse,ziyue2021tensor,li2022individualized,li2023tensor,li2023choose}, deep learning 
\cite{chen2023adaptive,xu2023kits,guo2024online}, contrastive learning\cite{ruan2023privacy,mao2022jointly,wang2023correlated,li2024non}, and Large Language Models (LLMs) \cite{liu2024spatial,liu2024timecma,zhang2024dualtime}. It is challenging to forecast \acrfull{NGD} due to the complex interplay of long-term trends, short-term seasonality, and uncertainties. 
Existing machine learning works utilize a structure-calibrated support vector regression (SVR) \cite{bai2016daily}, \acrshort{ANN} \cite{szoplik2015forecasting} or \acrfull{LSTM} Networks \cite{anagnostis2019applying} to forecast natural gas consumption. In \cite{gensler2016deep}, they integrated deep learning techniques to demonstrate their superior forecasting capabilities for energy output. Their study found that a hybrid model based on \acrshort{LSTM} networks achieved higher prediction accuracy than conventional ANN models like standard MLP and other physical forecasting models. This is also supported in \cite{wei2019daily}. A more extended comparison finds that \acrshort{LSTM} is superior to ARIMA, exponential smoothing, SVM, ANN, and recurrent neural network \cite{anagnostis2019applying,liu2023enhancement}. A novel approach is Prophet, an open-source algorithm developed by Facebook in 2017, which has been utilized in electricity price forecast \cite{stefenon2023aggregating}. A comparison study on monthly peak energy demand found this algorithm to perform best among SARIMA, LSTM, RNN and Prophet \cite{chaturvedi2022comparative}. In particular for the longer time windows, Prophet has been found to perform well \cite{taylor2018forecasting}.

\subsection{Impact of the Russia-Ukraine war on Natural Gas Demand}

A couple of research studies have been conducted to reckon the impact of the Russia-Ukraine war on Germany's future natural gas demand NGD across different sectors. In \cite{bachmann2022if}, the economic effects of a potential cut-off of Russian energy imports on the German economy are discussed. \cite{halser2022pathways} delve into the intricate dynamics of the natural gas crisis gripping Europe with a special focus on the German unique role, discussing strategies for mitigating the gas supply shortage. \cite{salzmannpredicting} applies time series models to forecast daily household demand, industrial demand with regards to gas-intensive production, and gas demand resulting from electricity generation of gas plants in Germany without explicitly assessing the impact of the Russia-Ukraine war on the model. \cite{ruhnau2023natural} examines the effects of the energy crisis on German NGD using an econometric model and found that gas consumption reductions were more pronounced in the industrial than the household sector. In contrast to this, our work is the first one to explicitly investigate the dependencies between various factors, including the war, and the natural gas demand, based on causal learning \cite{lan2023mm,lin2023dynamic,lan2024multifun}. It realizes this by including non-linear granger causality tests, which are not broadly applied yet.

\section{Methodology}
\label{sec:ra}

\subsection{Research Approach}
Three sectors of gas consumption will be considered separately, namely \acrfull{LDZ}, \acrfull{IND} and \acrfull{GTP}, which we believe will be affected by the conflict differently due to their nature. The aggregated natural gas demand is denoted as NGD. For example, 
\acrshort{LDZ} represents domestic and small commercial consumers, which has been found to be more influenced by temperature and holidays \cite{liu2018natural}. \acrshort{IND} is likely correlated with economic indicators. \acrshort{GTP} will likely depend strongly on the overall gas supply. As shown in Fig. \ref{fig:Researchapproach}, the process begins with selecting influencing factors that are determined through exploratory data analysis and mutual information analysis \cite{kraskov2004estimating}. The subsequent step involves applying deep learning in non-linear Granger-Causality tests to determine if the selected factors can indeed predict changes in (\acrshort{NGD}). These tests validate the capacity of selected predictors to provide a reliable forecast of \acrshort{NGD} and ensure that only the most relevant predictors are included in the model. Finally, the refined data are incorporated into the deep learning intervention studies to forecast \acrshort{NGD} in the context of the post-war scenario in Germany.

\begin{figure}
    \centering
    \includegraphics[width=\linewidth]{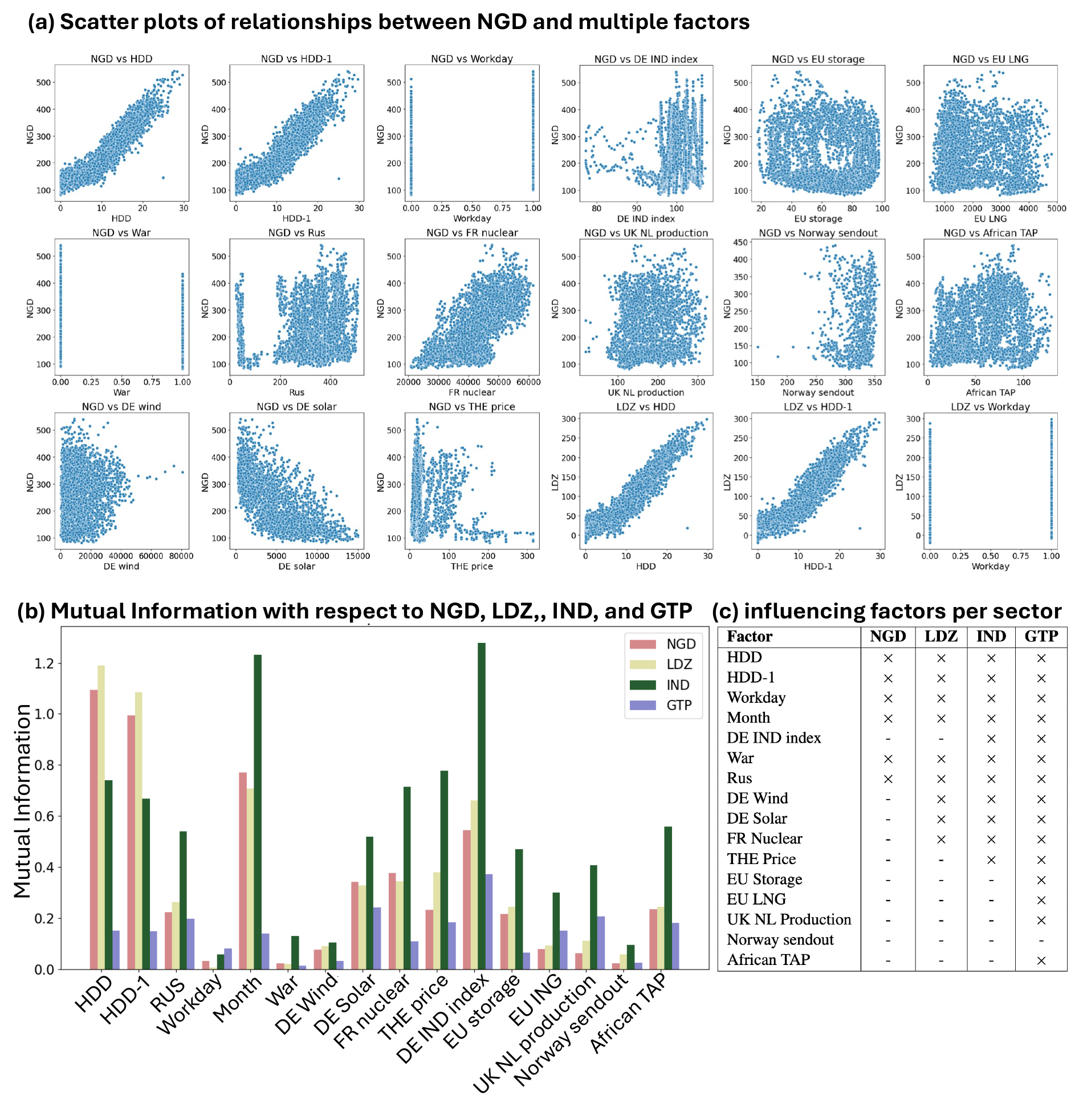}
    \captionof{figure}{(a) Scatter plots of relationships between NGD and multiple factors; (b) Mutual information analysis; (c) Selected factors (denoted with ``$\times$'') per sector and the aggregated NGD}
    \label{fig:Mutual information}
    \vspace{-10pt}
\end{figure}
\subsection{Data Sources and Definition}

The aggregated NGD data were retrieved from Bruegel, Bundesnetzagentur, and Trading Hub Europe, covering the period from April 2010 to July 2023. The factors for predicting gas demand are defined based on previous research. One part of the prediction is the previous gas demand. For temporal effects, the month and holidays are included into the predictive model. In addition, German natural gas consumption strongly depends on temperature \cite{rusev2021multiple}. Heating degree days (HDD) is a versatile measure of cold weather duration and severity in 24 hours compared to a base temperature, allowing for weather-based analysis of natural gas consumption \cite{meng2021evaluating}. In addition to a contemporary influence, the previous day's can also influence today's demand due to heating system dynamics and user behaviors \cite{demirel2012forecasting}. To control for the Russian-Ukraine's impact on NGD, a `War' dummy variable is instituted, with the February 24, 2022 as anchor. Next, the various gas supply sources are included, as well as the gas storage levels. Disruptions in supply can influence future demand, especially if there are pipeline outages or if users anticipate further supply issues \cite{berger2022potential}. Important outside predictors are the gas price, which can incorporate information unknown to the model, and industry production, which is another large consumer of gas in Germany. In systems where natural gas is used as a backup or supplementary source to renewable generation, the output from other energy sources can influence gas demand. To incorporate this, renewable energy production is also included in the model. In addition to that, cross country effects could play a role, especially to France, which had issues in its nuclear production capability. Thus, the french nuclear energy production is also included in the model.

\section{Implementation and Results}
\label{sec: result}
\subsection{Data Exploration and Mutual Information Analysis}

In the dataset, a number of dependencies are found. As shown in Fig. \ref{fig:Mutual information}(a), a pronounced linear relation between NGD and HDD suggests that colder temperatures elevate gas demand. This trend is consistent even when considering lagged HDD, reinforcing the direct linear relationship between temperature and gas demand. For the other variables, no linear relationship is found. Fig. \ref{fig:Mutual information}(b) provides a quantitative overview by mutual information analysis of the significant determinants influencing gas demand across various sectors. For aggregated demand (NGD), the primary driving forces are \acrshort{HDD}, lagged 1 \acrshort{HDD} (\textit{HDD-1}), Russian gas flow into Europe (\textit{RUS}), Workday patterns, and monthly variations. This is further supplemented by geopolitical events (War), as well as the production of renewable energy sources such as German wind energy and German solar energy. The \acrshort{LDZ} sector is broadly influenced by similar factors but has an added emphasis on French nuclear power production. \acrshort{IND} sector gas demand showcases a blend of climatic conditions (\textit{\acrshort{HDD}}, \textit{\acrshort{HDD}-1}), natural gas origins (\textit{Rus},\textit{ UK NL Production}), and \textit{THE Price}, alongside renewable energy inputs. Meanwhile, the GTP sector is majorly swayed by most of the influencing factors, ranging from monthly fluctuations in geopolitical events to a set of renewable energy sources and pricing mechanics. The selection of influencing factors on German NGD summarized below would capture approximately 90\% quantile of mutual information in each sector. The geopolitical event (the War) and natural gas import dynamics (\textit{Rus}) have been included in each sector to intricate interplay of geopolitical events in shaping the gas demand landscape across these sectors.

\begin{figure}
    \centering
    \includegraphics[width=0.99\linewidth]{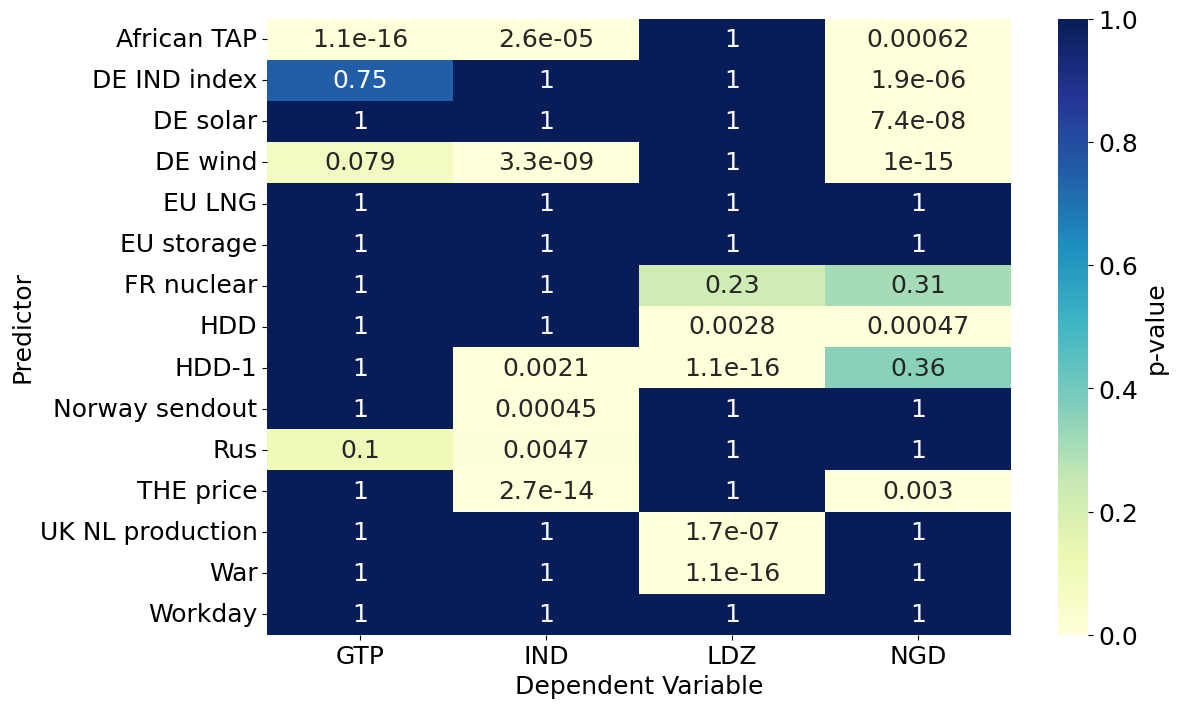}
    \caption{LSTM-based Granger Causality Test Result}
    \label{fig:granger}
    \vspace{-10pt}
\end{figure}

\subsection{Nonlinear Granger Causality}

\begin{figure}
    \centering
    \includegraphics[width=0.99\linewidth]{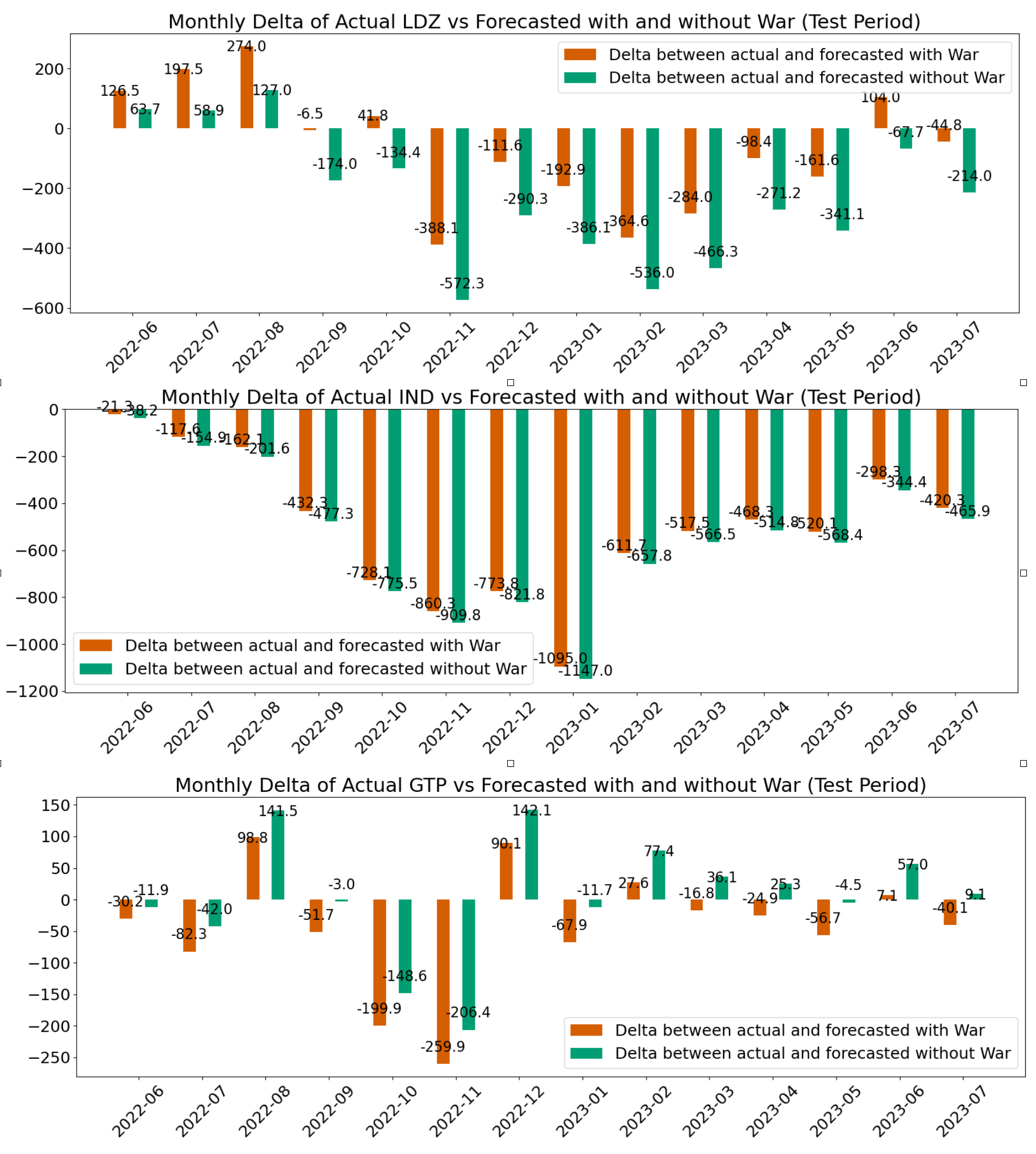}
    \caption{Monthly difference ($\Delta$)  (in mcm) of forecasts with (red) and without (green) the war comparing with actual demands: top (LDZ), mid (IND), and bottom (GTP).}
    \label{fig:Monthly delta}
    \vspace{-10pt}
\end{figure}

Granger causality is the popular choice to examine the causal relations among time-series. Standard approaches to Granger causality detection commonly assume linear dynamics, due to the nonlinearity, we adopted a non-linear Granger causality test based on LSTM to analyze the dependencies of gas demands. The key idea of Granger causality is: a time series $\mathbf{x}$ is said to be Granger-cause of another time series $\mathbf{y}$ if past values of $\mathbf{x}$ provide significant information about the future values of $\mathbf{y}$ beyond what is already provided by the past values of $\mathbf{y}$ itself. Thus, a statistical test will be conducted by comparing the following two:
\begin{equation}
\begin{split}
    & y_t = f_1(\mathbf{y}_{t-\tau : t-1}, \mathbf{Z}_{t-\tau : t-1}) + e\\
    & y'_t = f_1(\mathbf{y}_{t-\tau : t-1}, \mathbf{Z}_{t-\tau : t-1} + \mathbf{x}_{t-\tau : t-1}) + e'\\
    & x \text{ being the cause of } y = \textit{Wilcoxon-test}(y_t, y'_t)
\end{split}
\end{equation}
where for example, our $y, y'$ is each energy demand (GTP to NGD), $x$ is the each predictor in Fig. \ref{fig:granger}, the $\mathbf{Z}$ are the other control variables for prediction, $\tau$ is the lag length, $e$ is the error, and $f_1(\cdot)$ is the one-step prediction model, i.e., LSTM.

Since LSTM may not follow the assumed F-distribution, the Wilcoxon signed-rank test, a non-parametric test, is used to compare the two 
\cite{swarup2022effects}. The $p$-values are shown in Fig. \ref{fig:granger}: While \textit{HDD} only appears to have a significant non-linear Granger causal relationship with LDZ demand, \textit{\acrshort{HDD}-1} shows notable nonlinear Granger causality with IND as well as \acrshort{LDZ}. The gas price only has a direct influence on the industrial gas demand IND. Besides, a large proportion of predictors like \textit{EU storage}, \textit{EU LNG}, and \textit{Workday} consistently show \( p \)-values gravitate towards or equal 1 across all \acrshort{NGD} sectors which indicates an absence of non-linear Granger causality. These findings highlight the need to identify other key factors that can strongly predict outcomes in future research studies. Additionally, the predictor \textit{War} exhibits a significant non-linear Granger causality with  \acrshort{LDZ}, with an extremely small \( p \)-values of 1.1$\times 10^{-16}$.

We also use Prophet as a base model, i.e., $f(\cdot) = \text{Prophet}$. Due to space limits, we only conclude the common insights: for example, 
IND gas demand depends on \textit{DE wind} and \textit{THE price}. LDZ depends on \textit{HDD-1} and \textit{UK NL production}, and the overall NGD depends on \textit{DE index}, \textit{DE wind}, and \textit{HDD}.

\subsection{Counterfactual Intervention Studies}

\begin{figure}
    \centering
    \includegraphics[width=0.99\linewidth]{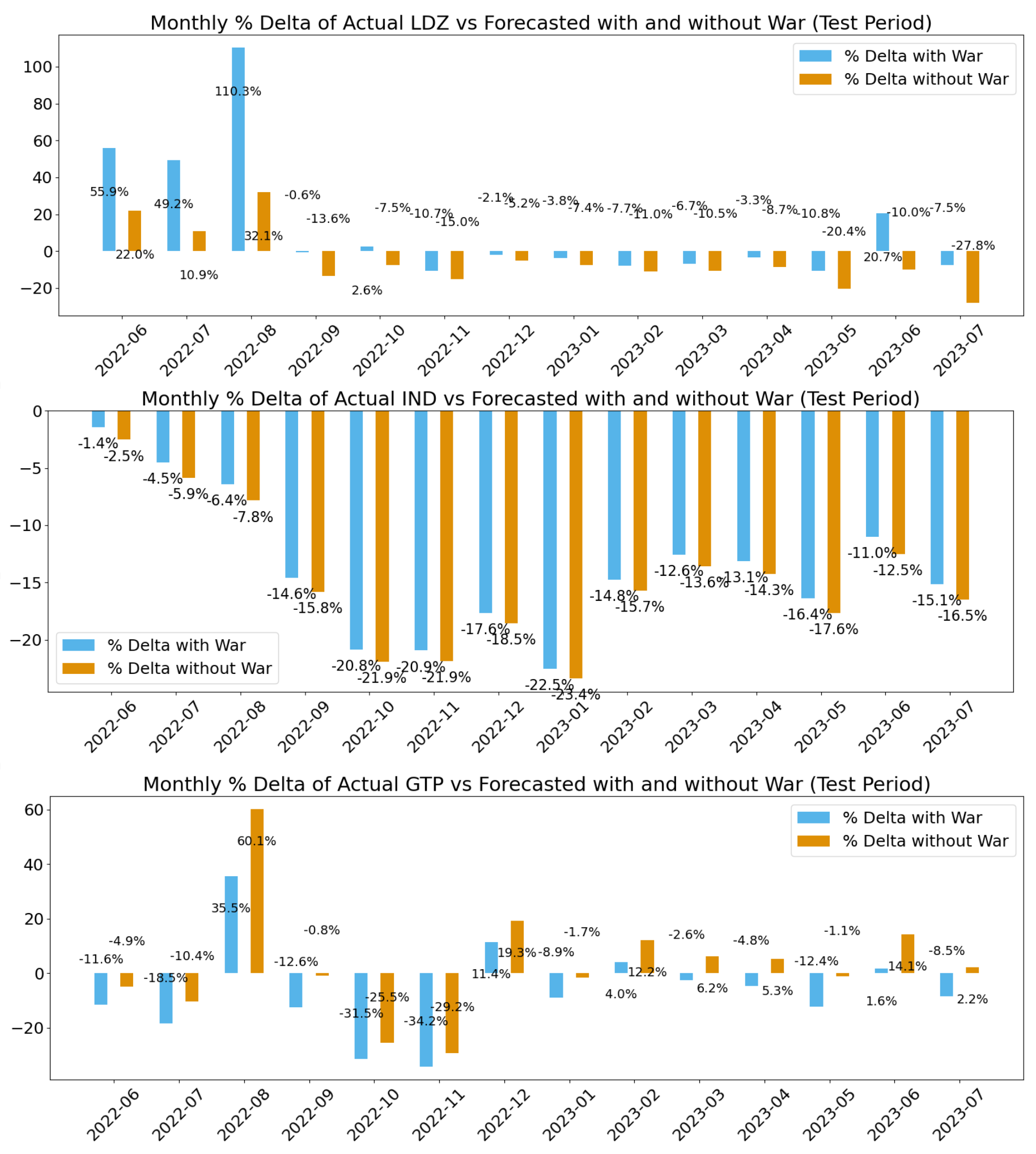}
    \caption{Monthly difference ($\Delta$)  (in percentage) of forecasts with (blue) and without (orange) the war comparing with actual demands: top (LDZ), mid (IND), and bottom (GTP).}
    \label{fig:delta in percentage}
    \vspace{-10pt}
\end{figure}

To explore the effect of the war on natural gas demand, a counterfactual intervention study is conducted. As a baseline, a prediction is taken without the data from the onset of the war, comparing it to the actual development.

\vspace{-10pt}
\begin{equation}
\begin{split}
    & y_t^{\textit{w.  war}} = f_1 (\mathbf{Z}_{t-\tau: t-1}, \mathbf{x}^{\textit{war}}_{t-\tau: t-1}) + e_1 \text{ (prediction)}\\
    & y_t^{\textit{w/o. war}} = f_{t-t_{\textit{war}}}(\mathbf{Z}_{t_{\textit{war}}-\tau: t_{\textit{war}}}) + e_2 \text{ (counterfactual)}\\
\end{split}
\end{equation}
where $\mathbf{Z}$ is all the other predicting factors except the war dummy variable, and $f_h(\cdot)$ is the $h$-step prediction model, and we used both LSTM and Prophet. To predict the case without war, only training data (without the war dummy variable) before the $t_{war}$ is used, but a multi-step prediction is needed, with $t - t_{war}$ steps. Otherwise, to predict the war case, all the historical data can be used, and additionally with a war dummy $\mathbf{x}^{\textit{war}}$.

Fig. \ref{fig:Monthly delta} presents sector-specific impacts in million cubic meters (mcm) per day of the intervention Russia-Ukraine War on German NGD. Fig. \ref{fig:delta in percentage} expresses deltas as percentages, meaning the relative deviation of forecasts from actual NGD. 

\textbf{In the LDZ sector}, the model without war shows a greater deviation from actual data than the model considering war, suggesting that war's impact was significant enough to be a determining factor in forecasting accuracy. The variation in delta values for the \acrshort{LDZ} sector indicates the impact of the war does not have a consistent pattern over the long term. Without considering the war dummy, the forecasted \acrshort{LDZ} seems closer to the actual data initially. From September 2022 onwards, the forecast considering the war aligns more closely with the actual \acrshort{LDZ} as time progresses, which suggests an increasing impact of the war on this sector.

\textbf{In the IND sector}, the impact of war appears to be more pronounced and consistent as the forecast demands incorporating the war are consistently closer to the actual figures. \textbf{This indicates that the war had an immediate and substantial effect on \acrshort{IND} sector and seems stable over time.} Since the deltas in the \acrshort{IND} sector are less volatile but consistently negative, the intervention led to a systematic decrease in \acrshort{NGD} compared to what was forecasted.

\textbf{In the GTP sector}, \textbf{the high deltas in the initial months suggest an immediate impact of the war.} However, the subsequent reduction and fluctuating deltas could imply that the obvious effects of the war diminish and become less predictable over time. The \acrshort{GTP} sector's response to the intervention fluctuates, suggesting that the impact does not maintain a consistent pattern over time and is influenced by additional factors beyond the war.

In summary, the impact of the Russia-Ukraine war as an intervention varies across German gas sectors. The \acrshort{LDZ} and \acrshort{GTP} sectors exhibited immediate responses with subsequent fluctuations, while the \acrshort{IND} sector showed a consistent trend and suggests a prolonged impact. The attribution of these effects to the intervention is most evident where the forecasts considering the war show distinct deviations from actual data, particularly in the early months of the intervention. Over the longer term, the persistence of these effects appears to be sector-dependent with sustained trends in \acrshort{IND} sector and more complex and unstable in \acrshort{GTP} sector.

\section{Discussion and Conclusion}
\label{sec: con}

The purpose of this paper is 
to integrate deep learning models within a causal analysis framework to evaluate the impact of the Russia-Ukraine conflict on German natural gas demand. 
The impact of the war on German natural gas demand is heterogeneously distributed across different gas sectors. 
The intervention study, constructing a counterfactual predicted time series hints at the capabilities could be derived from applying this approach in a broader scope. Even though the effect found here is rather small, it shows promising results in uncovering the causal effect of shocks using predictive methods.

We acknowledge the limitation in long-term forecasting accuracy 
since the prediction error will accumulate \cite{tadjer2021machine}. Despite this limitation, the adaptability of deep learning models remains as a strong asset 
to navigate and understand complex energy demand dynamics in times of geopolitical uncertainty.

\bibliographystyle{plain}
\bibliography{main}

\end{document}